\documentclass{article}

\usepackage{microtype}
\usepackage{graphicx}
\usepackage{subcaption}

\usepackage{hyperref}

\usepackage[accepted]{icml2026}

\usepackage{amsmath}
\usepackage{amssymb}
\usepackage{mathtools}
\usepackage{amsthm}
\usepackage{comment}

\usepackage[capitalize,noabbrev]{cleveref}

\theoremstyle{plain}

\theoremstyle{definition}

\theoremstyle{remark}

\usepackage[textsize=tiny]{todonotes}

\icmltitlerunning{GO-PRE: Goal-Oriented Next-Best-View Selection via Predictive Rendering Entropy for Active 3D Reconstruction}

\usepackage{booktabs}
\usepackage{multirow}
\usepackage{diagbox}
\usepackage[table]{xcolor}
\usepackage{colortbl}
\usepackage{enumitem}

\definecolor{Best}{RGB}{248,190,190}%
\definecolor{Second}{RGB}{252,220,185}%
\definecolor{Third}{RGB}{255,245,180}%

\newcommand{\best}[1]{\cellcolor{Best}{\textbf{#1}}}
\newcommand{\second}[1]{\cellcolor{Second}{#1}}
\newcommand{\third}[1]{\cellcolor{Third}{#1}}

\begin{document}

\twocolumn[
  \icmltitle{GO-PRE: Goal-Oriented Next-Best-View Selection via Predictive Rendering Entropy for Active 3D Reconstruction}

    \begin{icmlauthorlist}
    \icmlauthor{Yan Song}{fudan}
    \icmlauthor{Zhihao Li}{sdu}
    \icmlauthor{Chenglong Li}{sdu}
    \icmlauthor{Li He}{fudan}
    \icmlauthor{Yan Wang}{ecnu}
    \icmlauthor{Wenqiang Zhang}{fudan}
  \end{icmlauthorlist}

  \icmlaffiliation{fudan}{College of Intelligent Robotics and Advanced Manufacturing, Fudan University, Shanghai, China}
  \icmlaffiliation{sdu}{School of Software, Shandong University, Jinan, China}
  \icmlaffiliation{ecnu}{School of Data Science and Engineering, East China Normal University, Shanghai, China}

  \icmlcorrespondingauthor{Yan Wang}{yanwang@dase.ecnu.edu.cn}
  \icmlcorrespondingauthor{Wenqiang Zhang}{wqzhang@fudan.edu.cn}

  \icmlkeywords{Machine Learning, ICML}

  \vskip 0.3in
]

\printAffiliationsAndNotice{}%

\begin{abstract}
  Active 3D reconstruction relies on active view selection to maximize reconstruction fidelity under limited capture budgets. However, most existing methods rely on surrogate signals such as parameter uncertainty or geometric heuristics, but these signals are often misaligned with the ultimate goal: the fidelity of rendered predictions. We propose GO-PRE, a goal-oriented next-best-view selection framework that explicitly targets information gain in the prediction space. Specifically, we formulate the objective as maximizing the reduction of the average marginal predictive entropy over a user-specified target view manifold. GO-PRE supports interactive goal specification and yields an efficient acquisition rule that enables real-time computation of information gain. Extensive experiments across benchmarks demonstrate that GO-PRE consistently improves active reconstruction performance and provides more reliable uncertainty quantification compared to state-of-the-art methods.

\end{abstract}

\section{Introduction}

Differentiable rendering techniques, such as Neural Radiance Fields~\cite{mildenhall2021nerf} and 3D Gaussian Splatting~\cite{kerbl3Dgaussians}, have enabled high-fidelity novel view synthesis and 3D reconstruction. However, in real-world capture and robotic exploration, the number of images is constrained by time, energy, and navigational reachability. This motivates active 3D reconstruction with next-best-view  selection~\cite{chen2024gennbv,ye2024pvp}, which iteratively chooses the next camera pose to maximize information gain under a limited budget.
\begin{figure}[t]

    \centering
    \includegraphics[width=1\linewidth]{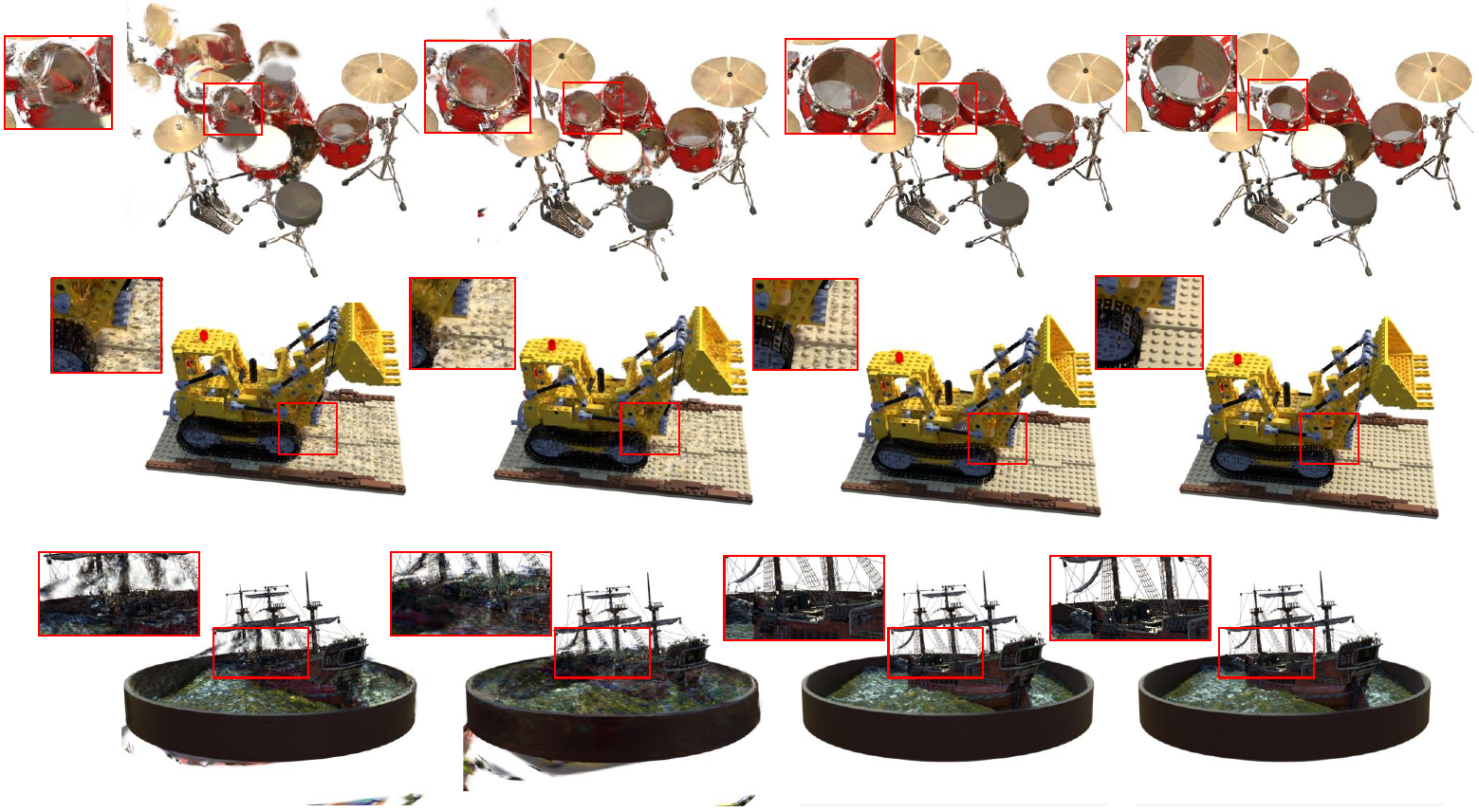}
    \vspace{1mm}
    \hspace*{0.05\linewidth}%
    \begin{tabular}{@{}cccc@{}}
      \small (a) Random & \small (b) FisherRF & \small (c) Ours & \small (d) Ground Truth
    \end{tabular}
    \caption{We propose GO-PRE, a novel framework for quantifying information gain in active 3D reconstruction. By directly targeting the reduction of rendering entropy in the prediction space, we ensure that the selected views maximize visual fidelity. In the above figure, each method is provided with a pool of one hundred candidate views from the Blender dataset and selects ten views to train a 3D Gaussian Splatting model. Compared to state-of-the-art approaches, our method more accurately estimates the information value of images, leading to significantly improved reconstruction of complex geometries and textures.}
    \label{fig:teaser}
\end{figure}

A key challenge is how to quantify information gain. Many existing next-best-view strategies~\cite{jiang2023fisherrf,xie2025gauss,wilson2025pop} rely on surrogate signals such as parameter-space uncertainty or geometric heuristics~\cite{jin2025activegs,chen2025activegamer}, which do not directly reflect the quantity that ultimately matters in differentiable reconstruction: the accuracy of unseen renderings. As a result, a method may reduce parameter variance while leaving rendering artifacts unresolved, leading to inefficient view selection and loss of fine details.

To bridge this misalignment and to enable controllable reconstruction, we propose GO-PRE, a goal-oriented next best view selection framework on the 3D Gaussian Splatting~\cite{kerbl3Dgaussians} backend, directly driven by predictive rendering entropy. Our central premise is that information gain for view selection should be defined in a principled manner in prediction space---namely, by targeting the \emph{expected reduction in uncertainty of novel-view synthesis within the target manifold~\cite{lindley1956measure,6795562}}, thereby naturally aligning view selection with the rendering quality that ultimately matters. To this end, we use a user-specified target view manifold to represent the user's viewpoints of interest. For any target pose within the manifold, we treat its rendering as the quantity of interest and measure uncertainty at that pose via the marginal predictive entropy ~\cite{Shannon1948,Houlsby2011}. We further define the overall uncertainty as the \emph{average marginal predictive entropy} over the target view manifold, and formulate the next-best-view objective in a mutual-information style as \emph{expected uncertainty reduction}: selecting the next view that maximizes the expected decrease in the predictive uncertainty over the target view manifold after acquiring a new observation. In this way, view selection is no longer driven by indirect proxies such as parameter variance or geometric heuristics, but by an interpretable objective that is directly aligned with novel-view synthesis quality.

Beyond aligning view selection with visual fidelity, this principled formulation naturally extends to support interactive goal specification. Users can specify target view manifold---augmented with a small number of probes---as a lightweight way to express ``where to focus / what to evaluate / which task matters,'' enabling the same reconstruction system to switch controllably among strategies such as global exploration, local refinement, and task-view optimization. Because the next-best-view objective acts directly on predictive uncertainty over the target view manifold, this interaction is not merely a preference setting, but a structured constraint on information gain: the system prioritizes views that most effectively reduce rendering uncertainty in the target view distribution, thereby better matching the resource allocation logic of real applications~\cite{8461098}. We conduct extensive evaluations on multiple benchmarks~\cite{knapitsch2017tanks,mildenhall2021nerf,barron2022mip}; both quantitative and qualitative results clearly demonstrate significant advantages over prior methods and heuristic baselines, while providing explicit interpretability.

To summarize, our main contributions are as follows:
\begin{itemize}[leftmargin=*,noitemsep,topsep=0pt]
    \item A principled next-best-view objective based on the reduction of \emph{average marginal predictive entropy} to align view selection with rendering quality.
    \item A goal-oriented framework enabling interactive goal specification and controllable strategy switching via the target view distribution.
    \item An efficient and highly interpretable method for predictive rendering uncertainty estimation and visualization.
    \item Extensive comparative studies demonstrating that our approach outperforms state-of-the-art methods in both view selection and uncertainty prediction.
\end{itemize}
\begin{figure*}[t]
  \centering
  \includegraphics[width=\textwidth]{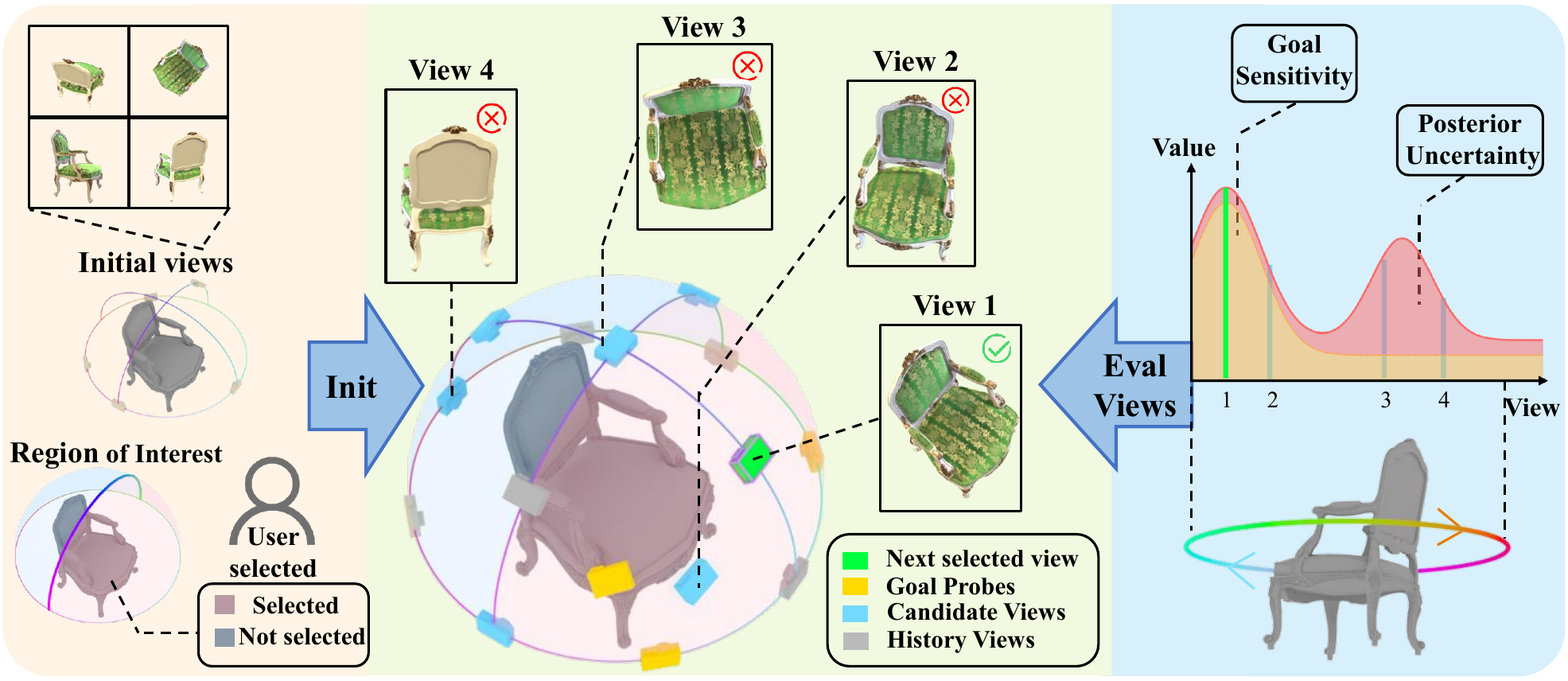}
  \caption{\textbf{Overview of the GO-PRE pipeline}. 
  The process begins with interactive target view manifold (left), where a user defines a Region of Interest. Probes (yellow) are then sampled to quantify the goal sensitivity of the scene parameters to this target view manifold. The decision mechanism (right) illustrates the core advantage of our approach: although View 3 exhibits high global Posterior Uncertainty (red shaded area), it is rejected because it contributes little to the target task. In contrast, View 1 is chosen as the Next Selected View (green line) because it aligns with the peak Goal Sensitivity (orange curve), thereby maximizing the expected reduction of predictive entropy within the user-specified target view manifold.}
  \label{fig:method-overview}
\end{figure*}

\section{Related Work}

\subsection{Uncertainty-aware 3D Representations}
While radiance fields enable photorealistic novel view synthesis, standard formulations are deterministic and lack intrinsic confidence measures. For NeRF, early efforts capture epistemic uncertainty via density-aware ensembles~\cite{sunderhauf2022density} or Bayesian variational inference~\cite{shen2021stochastic}. Another line of work models radiance as a random variable, using the posterior color variance to quantify uncertainty in ActiveNeRF~\cite{pan2022activenerf} and a variance-based PSNR proxy in NeurAR~\cite{ran2023neurar}. With the advent of 3D Gaussian Splatting~\cite{kerbl3Dgaussians}, research has shifted toward quantifying uncertainty over discrete primitives, either by modeling Gaussian parameters as probabilistic variables~\cite{li2024variational,lyu2024manifold} or by attaching auxiliary uncertainty attributes to primitives~\cite{han2025view,tan2025uncertainty}.

A more direct approach, closer to our work, investigates prediction-space statistics: Ewen et al.~\cite{ewen2025these} derive differentiable higher-order moments of the rendering equation, while Gottwald et al.~\cite{gottwald2025primu} splat primitive-level training errors and visibility onto novel views and regress them against holdout data. In contrast, our predictive uncertainty is epistemic by construction—the entropy of a Gaussian induced by a Laplace posterior under a linearized renderer—not a deterministic descriptor of the trained model.

\subsection{Active 3D Reconstruction}
Active reconstruction is a classical problem in computer vision and robotics~\cite{chen2011active,lluvia2021active}, aiming to autonomously select the next-best view that maximizes reconstruction fidelity under limited acquisition costs. Traditional methods reason over volumetric representations such as OctoMap~\cite{hornung2013octomap} and exploit geometric proxies for information gain, most notably frontiers between mapped and unmapped regions~\cite{cieslewski2017rapid,zhou2021fuel} or under-observed surface boundaries inferred directly from point clouds~\cite{border2024surface}. Data-driven approaches instead learn the decision rule, via reinforcement-learning policies~\cite{chen2024gennbv}, learned uncertainty distributions~\cite{shen2025auto3r}, or regressed image-quality scores~\cite{wang2025active}; they rely on learned predictors and can be sensitive to training distributions. Neu-NBV~\cite{jin2023neu} performs mapless NBV planning by estimating per-pixel uncertainty in image-based neural rendering. A concurrent work partitions the Gaussian model into consistent regions and selects the next informative pose via the semantic feature variance of Gaussians, while additionally handling pose noise \citep{li2026active}. To provide a rigorous theoretical foundation, recent works frame view selection as information-gain maximization in the spirit of Bayesian experimental design~\cite{lindley1956measure}. For radiance fields, FisherRF~\cite{jiang2023fisherrf} uses the Fisher Information to maximize the Expected Information Gain over model parameters; POp-GS~\cite{wilson2025pop} recasts this through optimal experimental design with a P-Optimality family and a block-diagonal covariance approximation that captures parameter correlations; and GauSS-MI~\cite{xie2025gauss} instead quantifies the Shannon mutual information~\cite{julian2014mutual, zhang2020fsmi} of per-Gaussian visual uncertainty under a candidate viewpoint. Unlike these parameter-space or surrogate criteria, GO-PRE directly minimizes the Average Marginal Predictive Entropy over a target view manifold, aligning view selection with rendering fidelity itself.

\section{Method}

This section presents the probabilistic formulation of goal-oriented predictive rendering, followed by the derivation of our next-best-view objective based on expected information gain in the prediction space.

\subsection{Probabilistic Modeling of Predictive Uncertainty}

To ensure consistency between the active reconstruction objective and actual visual fidelity, we formulate information gain directly within the predictive rendering space. We introduce the Target View Manifold $\mathcal{M}$---a continuous region in the camera pose space---to represent the user's viewpoints of interest. To formulate the information gain, we define $q(\cdot)$ as a distribution supported on $\mathcal{M}$, and establish the active reconstruction objective as maximizing the predictive rendering information gain over this manifold.

For any pose $p$ within the target view manifold, we model the predictive rendering $z(p)$ as
\begin{equation}
z(p) = f(p, \theta) + \eta,
\quad
\eta \sim \mathcal{N}(0, \tau^2 I),
\label{eq:predictive-model}
\end{equation}
where $f(p, \theta)$ denotes the differentiable rendering model parameterized by the scene parameters $\theta$, which are optimized using the currently collected observations $D$. The noise term $\eta$ represents inherent observation noise, capturing aleatoric uncertainty and ensuring that the predictive distribution remains non-degenerate.

The rendering uncertainty over the target view manifold is denoted as $\mathcal{U}(D)$ and quantified by the expected marginal predictive entropy:
\begin{equation}
\mathcal{U}(D) =
\mathbb{E}_{p \sim q}
\big[
H\big(z(p) \mid D\big)
\big].
\label{eq:global-uncertainty}
\end{equation}

Following principles from Bayesian experimental design, our view selection strategy aims to identify the candidate view $x^\star$ from a candidate pool $\mathcal{X}_{\text{pool}}$ that maximizes the expected information gain. Equivalently, this corresponds to minimizing the expected posterior uncertainty after acquiring a new observation $y_x$:
\begin{equation}
x^\star =
\arg\min_{x \in \mathcal{X}{\text{pool}}}
\mathbb{E}{y_x}
\big[
\mathcal{U}\big(D \cup {(x, y_x)}\big)
\big].
\label{eq:nbv-objective}
\end{equation}

\subsection{Deriving the GO-PRE Score via the Goal 
Information Matrix}
\label{sec:go-pre-score}

\begin{figure}
    \centering
    \includegraphics[width=1\linewidth]{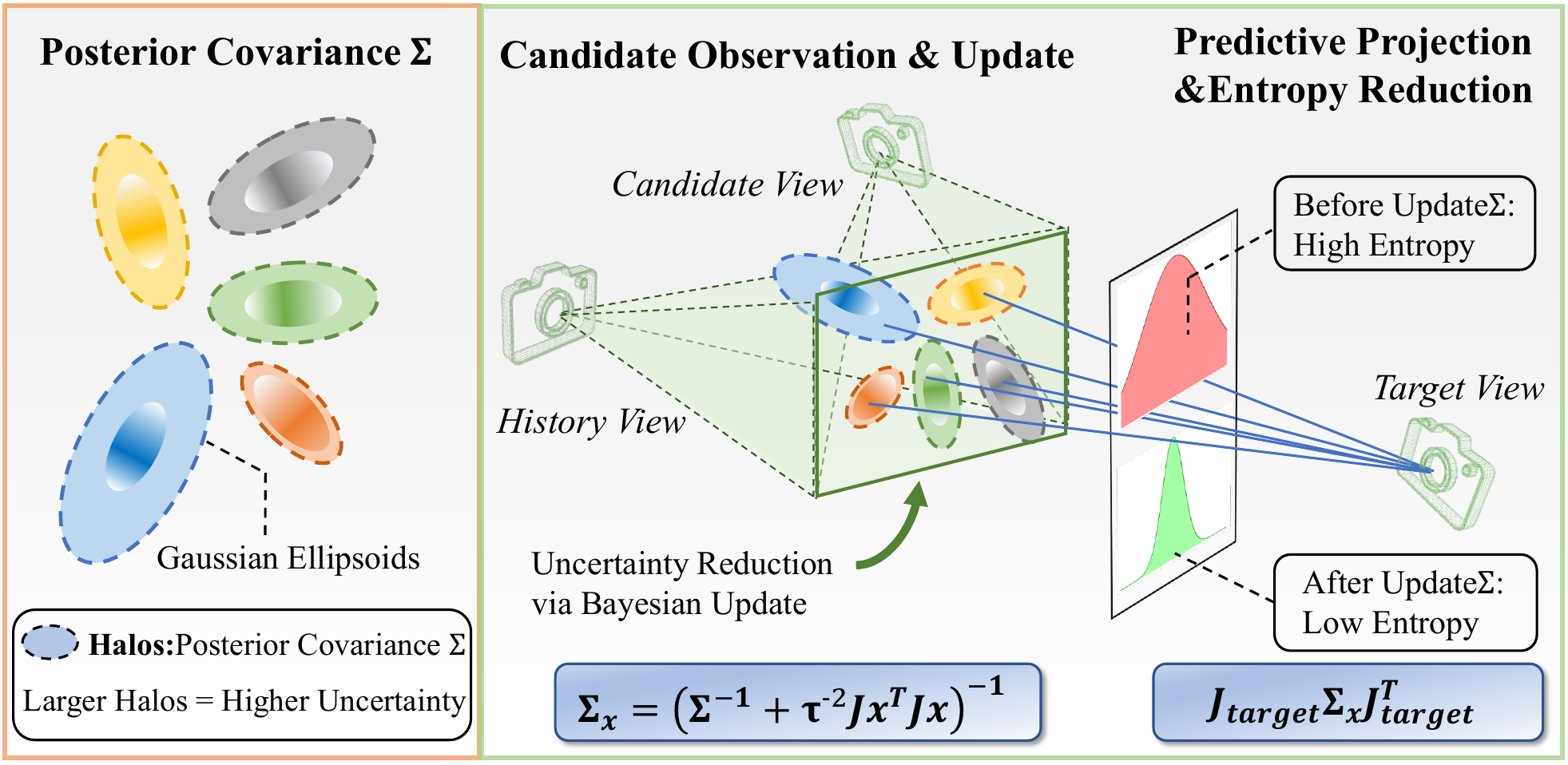}
    \caption{\textbf{Entropy-based View Selection.}
The framework connects parameter space to prediction space. Starting with the current Posterior Covariance, we simulate the Bayesian Update $\Sigma \to \Sigma_x$ derived from a candidate view. Finally, we project this update to the target view to compute the Predictive Entropy Reduction. The system prioritizes candidates that maximally tighten the predictive distribution, transitioning from Red distribution  to Green distribution}
    \label{fig:entropy-view-selection}
\end{figure}

We approximate the posterior distribution $p(\theta \mid D)$ via a Laplace approximation centered at the MAP estimate $\theta^*$ of the current model parameters, with covariance $\Sigma$. Accordingly, the renderer $f(p, \theta)$ is linearized using the Jacobian $J_p = \partial f(p, \theta) / \partial \theta |_{\theta^*}$. Under this linearization, the marginal predictive distribution $p(z(p) \mid D)$ becomes a Gaussian. The predictive uncertainty in Eq.~\ref{eq:global-uncertainty} simplifies to the expectation of the log-determinant of the predictive covariance:
\begin{equation}
\mathcal{U}(D) \equiv \mathbb{E}_{p \sim q} \left[ \frac{1}{2} \log \det \left( J_p \Sigma J_p^\top + \tau^2 I \right) \right].
\label{eq:global-entropy}
\end{equation}
As depicted in Figure~\ref{fig:entropy-view-selection}, under this linear-Gaussian formulation, observing a candidate view $x$ updates the parameter covariance $\Sigma$ to $\Sigma_x$ via a standard precision-additive update:
\begin{equation}
\Sigma_x = \left( \Sigma^{-1} + \tau^{-2} J_x^\top J_x \right)^{-1}.
\label{eq:covariance-update}
\end{equation}
This yields the predictive uncertainty: 
\begin{equation} \mathcal{U}_x = \mathbb{E}_{p \sim q} \left[ \frac{1}{2} \log \det (J_p \Sigma_x J_p^\top + \tau^2 I) \right]. \label{eq:posterior-uncertainty} \end{equation} By applying the Matrix Determinant Lemma, we shift the predictive entropy computation from the output space to the parameter space (detailed in the Appendix~\ref{sec:appendix-derivations}):
\begin{equation}
\mathcal{U}_x \equiv \mathbb{E}_{p \sim q} \left[ \frac{1}{2} \log \det \left( I + \tau^{-2} \Sigma_x^{1/2} S(p) \Sigma_x^{1/2} \right) \right],
\label{eq:parameter-space-uncertainty}
\end{equation}
where $S(p) = J_{p}^\top J_{p}$ denotes the Fisher information of the target pose $p$, and $\tau^{2}$ represents the RGB measurement covariance that we set equal to one.

\begin{figure}[t]
    \centering
    \includegraphics[width=1\linewidth]{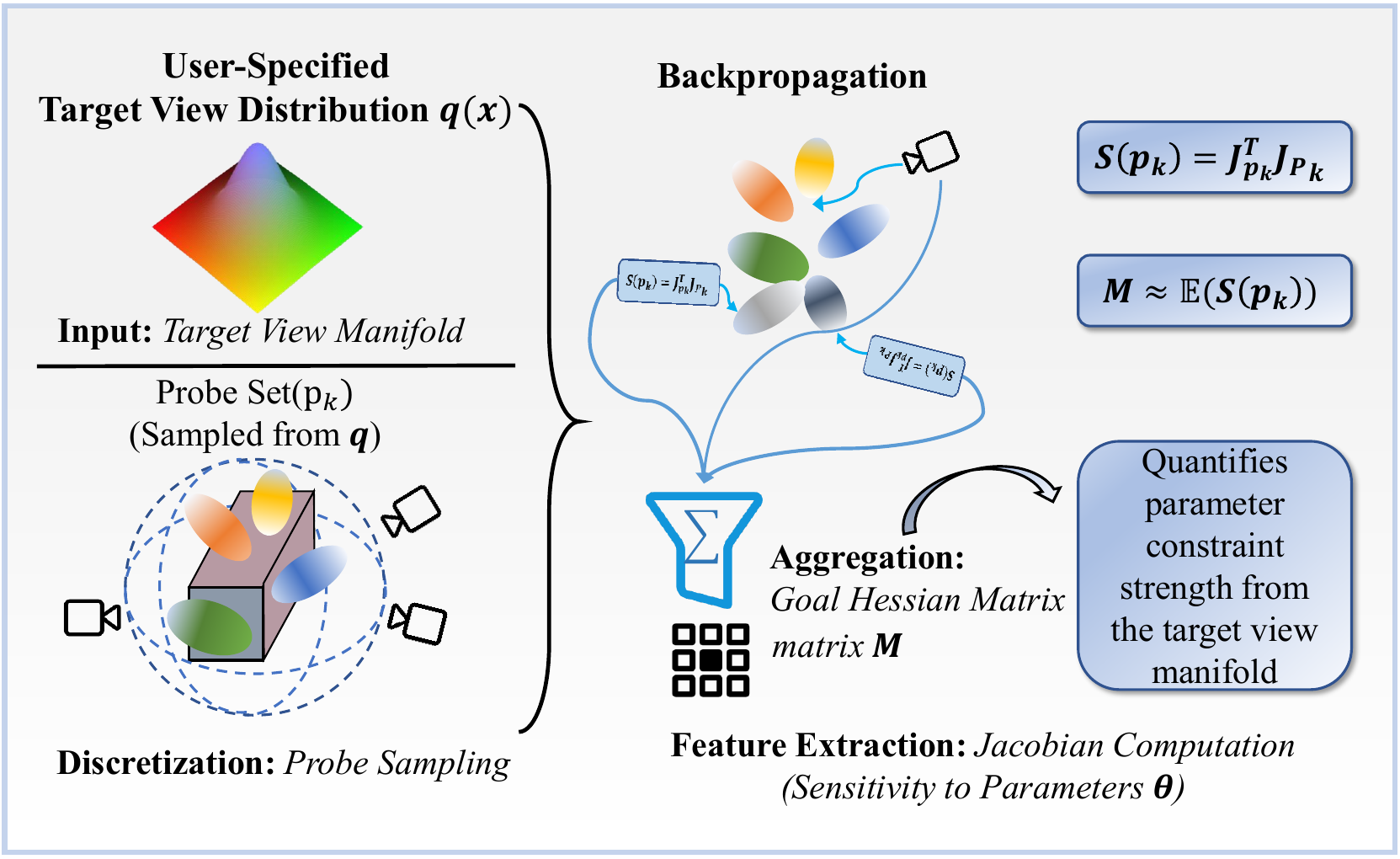}
    \caption{\textbf{Computation of the Goal Hessian Matrix $M$}. To quantify parameter importance relative to a user-defined goal, we sample discrete Probe Sets $\{p_k\}$ from the target view distribution $q(x)$. Through backpropagation, we compute the Fisher Information term $S(p_k) = J_{p_k}^T J_{p_k}$ for each probe. These terms are aggregated to form the Goal Hessian Matrix $M$, which serves as a compact representation of the "Goal Sensitivity," directing the reconstruction focus toward task-relevant regions.}
    \label{fig:goal-hessian}
\end{figure}

To enable efficient optimization, we derive a tractable surrogate objective by applying Jensen's inequality. Since the log-determinant is concave, swapping the expectation and the function evaluation yields a rigorous upper bound on the uncertainty, which we define as the GO-PRE Score:
\begin{equation}
\text{Score}_{\text{GO-PRE}}(x) = \log \det \big( I + \Sigma_x^{1/2} \mathbf{M} \Sigma_x^{1/2} \big).
\label{eq:go-pre-score}
\end{equation}
Here, $\mathbf{M}$ is the Goal Hessian Matrix. As illustrated in Figure~\ref{fig:goal-hessian}, since analytical computation of the expectation over the continuous distribution $q(\cdot)$ is intractable, we approximate $\mathbf{M}$ via Monte Carlo sampling using $K$ probe poses $\{p_k\}$ drawn from $q(\cdot)$:
\begin{equation}
\mathbf{M} = \mathbb{E}_{p \sim q} \big[ S(p) \big] \approx \frac{1}{K} \sum_{k=1}^K J_{p_k}^\top J_{p_k}, \quad \{p_k\} \sim q(\cdot).
\label{eq:goal-hessian}
\end{equation}
Combining these components, the final next-best-view selection criterion is given by:
\begin{equation}
x^\star = \mathop{\arg\min}_{x \in \mathcal{X}_{\text{pool}}} \log \det \big( I + \Sigma_x^{1/2} \mathbf{M} \Sigma_x^{1/2} \big).
\label{eq:final-objective}
\end{equation}

Although Eq.~(\ref{eq:final-objective}) enables information gain estimation without ground truth observations, the high dimensionality of 3DGS parameters makes direct determinant computation prohibitive. Owing to the spatial locality of 3DGS, the matrices $\Sigma_x$ and $\mathbf{M}$ contain only a limited number of off-diagonal elements. Following the methodology of FisherRF~\cite{jiang2023fisherrf}, we employ a diagonal approximation for both $\Sigma_x$ and $\mathbf{M}$. This simplifies the log-determinant into a scalable summation of scalar logarithms, ensuring real-time computation even in large-scale scenes.

\begin{figure*}[t]
  \centering
  \includegraphics[width=\textwidth]{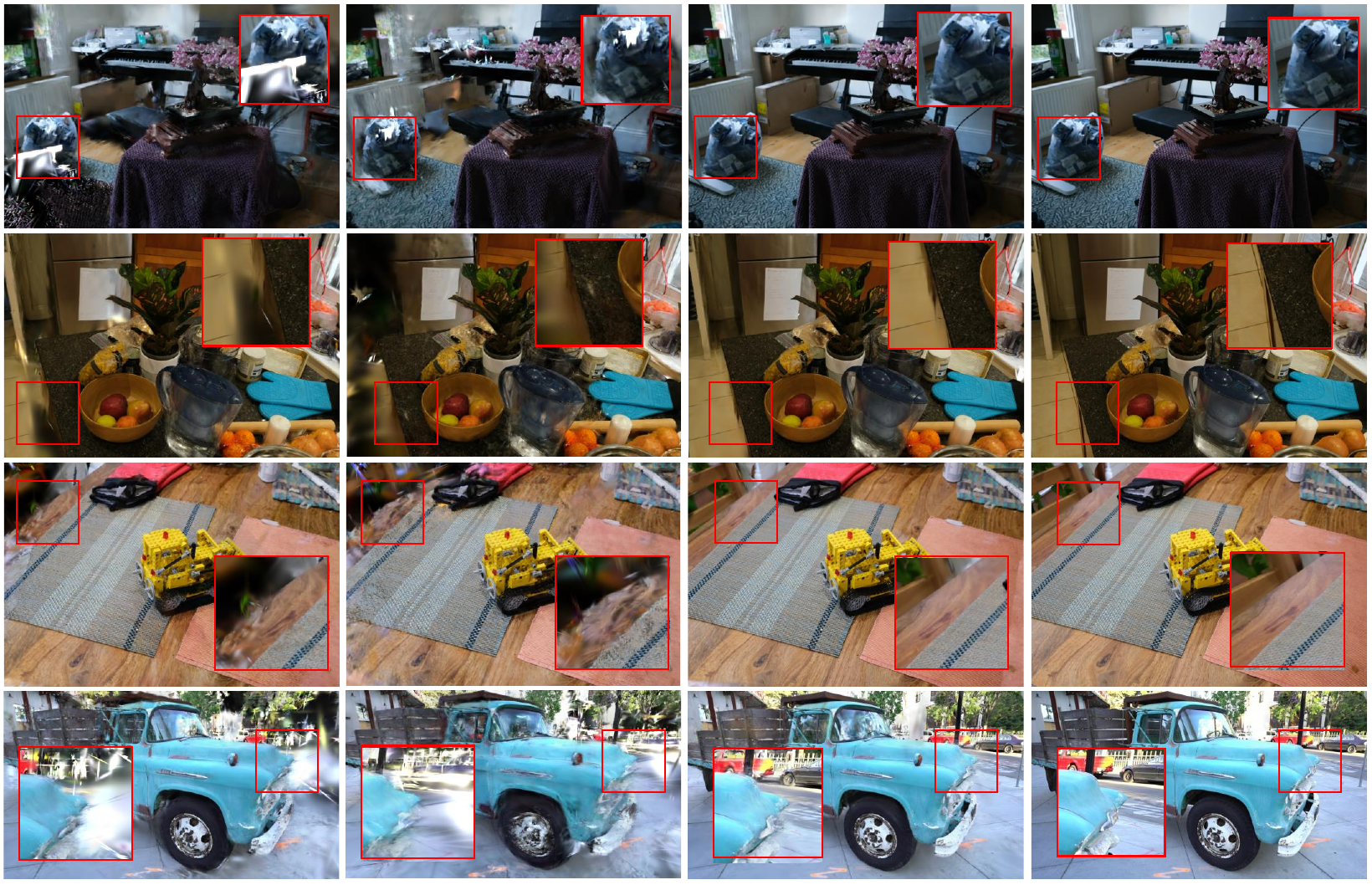}
   \makebox[0.23\textwidth][c]{\small (a) Random} 
    \makebox[0.23\textwidth][c]{\small (b) FisherRF }
    \makebox[0.23\textwidth][c]{\small (c) Ours}
    \makebox[0.23\textwidth][c]{\small (d) Ground Truth}
  \caption{ Qualitative comparisons on Mip-NeRF360 and Tanks 
  \& Temples datasets with 20 views}
  \label{fig:qualitative-results}
\end{figure*}

\subsection{Uncertainty Quantification}
\label{sec:uncertainty-quantification}
Beyond guiding the next-best-view selection, our probabilistic framework naturally enables the quantification of rendering uncertainty for any arbitrary viewpoint $x_0$. This capability is crucial for identifying under-reconstructed regions and providing a measure of reliability for the synthesized views. Based on the linearized predictive model established in Sec.~\ref{sec:go-pre-score}, the predictive distribution $p(z(x_0) \mid D)$ at a query pose $x_0$ follows a Gaussian distribution with covariance:\begin{equation}\Lambda_{x_0} = J_{x_0} \Sigma J_{x_0}^\top + \tau^2 I.\label{eq:predictive-covariance}\end{equation}Consequently, the predictive uncertainty is quantified by the differential entropy of this distribution:\begin{equation}H(z(x_0) \mid D) \equiv \frac{1}{2} \log \det (\Lambda_{x_0})\label{eq:predictive-entropy}\end{equation}where $J_{x_0} = \partial f(x_0, \theta) / \partial \theta |_{\theta^*}$ is the Jacobian evaluated at the query pose.
Eq.~(\ref{eq:predictive-entropy}) thus serves as a mathematically grounded metric to visualize the spatial distribution of reconstruction confidence. Unlike heuristic proxies, this predictive entropy directly reflects the informative value of the rendering, allowing for intuitive, real-time monitoring of the active reconstruction progress.

\begin{table*}[t]
\caption{Comparison on Mip-NeRF360, Tanks \& Temples and Blender. Higher PSNR/SSIM is better, lower LPIPS is better.}
\label{tab:global-reconstruction}
\centering
\setlength{\tabcolsep}{5pt}
\renewcommand{\arraystretch}{1.15}

\begin{tabular}{c|ccc|ccc|ccc}
\toprule
\multirow{2}{*}{\diagbox[width=9em]{Method}{Metrics}} &
\multicolumn{3}{c|}{Mip-NeRF360} &
\multicolumn{3}{c|}{Tanks \& Temples} &
\multicolumn{3}{c}{Blender} \\
\cline{2-10}
& PSNR$\uparrow$ & SSIM$\uparrow$ & LPIPS$\downarrow$
& PSNR$\uparrow$ & SSIM$\uparrow$ & LPIPS$\downarrow$
& PSNR$\uparrow$ & SSIM$\uparrow$ & LPIPS$\downarrow$ \\
\midrule

Random
& 17.9140 & 0.5640 & 0.4300
& 15.6210 & 0.5788 & 0.3876
& 22.4930 & 0.8730 & 0.1120 \\

ActiveNeRF
& 17.8890 & 0.5330 & 0.4140
& 15.8990 & 0.5872 & 0.3721
& 22.9790 & 0.8760 & 0.1110 \\

FisherRF
& \third{20.3510} & \third{0.6010} & \third{0.3610}
& \third{17.0133} & 0.6205 & 0.3445
& 23.6810 & 0.8830 & 0.1020 \\

GauSS-MI
& 19.8573 & 0.5835 & 0.3718
& 16.7400 & \third{0.6249} & \third{0.3415}
& \third{24.9826} & \third{0.8981} & \third{0.0882} \\

POp-GS
& \second{20.6180} & \second{0.6155} & \second{0.3505}
& \second{17.4055} & \second{0.6500} & \second{0.3185}
& \second{25.5235} & \second{0.9020} & \second{0.0805} \\

Ours
& \best{21.2283} & \best{0.6318} & \best{0.3370}
& \best{17.9570} & \best{0.6685} & \best{0.3045}
& \best{25.5740} & \best{0.9037} & \best{0.0800} \\

\bottomrule
\end{tabular}
\end{table*}

\begin{table*}[t]
\caption{Goal-Oriented results on Mip-NeRF360, Tanks \& Temples and Blender. Higher PSNR/SSIM is better, lower LPIPS is better.}
\label{tab:goal-oriented-reconstruction}
\centering
\setlength{\tabcolsep}{5pt}
\renewcommand{\arraystretch}{1.15}

\begin{tabular}{c|ccc|ccc|ccc}
\toprule
\multirow{2}{*}{\diagbox[width=9em]{Method}{Metrics}} &
\multicolumn{3}{c|}{Mip-NeRF360} &
\multicolumn{3}{c|}{Tanks \& Temples} &
\multicolumn{3}{c}{Blender} \\
\cline{2-10}
& PSNR$\uparrow$ & SSIM$\uparrow$ & LPIPS$\downarrow$
& PSNR$\uparrow$ & SSIM$\uparrow$ & LPIPS$\downarrow$
& PSNR$\uparrow$ & SSIM$\uparrow$ & LPIPS$\downarrow$ \\
\midrule

Random
& 18.4510 & 0.5550 & 0.3975
& 16.5935 & 0.6295 & 0.3560
& 23.0396 & 0.8745 & 0.1090 \\

FisherRF
& 19.8190 & 0.5806 & 0.3723
& 15.7124 & 0.5934 & 0.3877
& \third{25.3538} & \third{0.9007} & \third{0.0834} \\

GauSS-MI
& \third{19.9473} & \third{0.5892} & \third{0.3662}
& \third{16.6660} & \third{0.6370} & \third{0.3438}
& 25.3237 & 0.8983 & 0.0859 \\

POp-GS
& \second{20.6163} & \second{0.6075} & \second{0.3533}
& \second{16.7581} & \second{0.6491} & \second{0.3330}
& \second{26.1589} & \second{0.9049} & \second{0.0783} \\

Ours
& \best{23.4250} & \best{0.7035} & \best{0.2780}
& \best{20.5030} & \best{0.7730} & \best{0.2200}
& \best{27.5093} & \best{0.9173} & \best{0.0699} \\

\bottomrule
\end{tabular}
\end{table*}

\section{Experiments}
In this section, we conduct a comprehensive empirical evaluation of GO-PRE to validate its effectiveness in both global and goal-oriented active reconstruction. Our experiments are designed to assess three core aspects: (i) the efficiency of our predictive rendering entropy-based objective in guiding view selection across diverse benchmarks; (ii) the controllability of the framework in directing information acquisition toward user-specified target manifolds; and (iii) the reliability of our probabilistic formulation for quantifying rendering uncertainty. We further perform ablation studies to isolate the contributions of key components to the overall performance.
\subsection{Active View Selection}
\label{sec:active-view-selection}
We conducted extensive experiments to demonstrate that our predictive rendering entropy-based objective effectively guides the model to select the most informative views. Furthermore, we assess the framework's capability in goal-oriented reconstruction scenarios, validating its effectiveness in prioritization within user-specified target view manifold. Below, we detail the datasets, baselines, and experimental configurations used in our study.

\noindent\textbf{Datasets.} Our approach is extensively evaluated on three common benchmark datasets: the Blender~\cite{mildenhall2021nerf} dataset, the Mip-NeRF360~\cite{barron2022mip} dataset, and the Tanks \& Temples~\cite{knapitsch2017tanks} dataset. The Blender~\cite{mildenhall2021nerf} dataset comprises eight synthetic objects with intricate geometry and realistic non-Lambertian materials. The Mip-NeRF360~\cite{barron2022mip} dataset comprises 9 scenes, consisting of five outdoor scenes and four indoor scenes. This high-resolution real-world dataset is widely established as a standard benchmark for assessing novel view synthesis quality~\cite{barron2023zip,li20253d}. We train our models at a resolution of $1066 \times 1600$, following the configuration used in 3D Gaussian Splatting~\cite{kerbl3Dgaussians}. Additionally, we evaluate on the \textit{Train} and \textit{Truck} scenes from the Tanks \& Temples dataset. These scenes feature drone-style outdoor captures characterized by long-baseline parallax and strong depth discontinuities, allowing us to test the robustness of our method under extreme geometric challenges.

\noindent\textbf{Metrics.} We quantify the reconstruction fidelity using Peak Signal-to-Noise Ratio (PSNR) and Structural Similarity (SSIM)~\cite{wang2004image}. To account for high-level visual features, we also include Learned Perceptual Image Patch Similarity (LPIPS)~\cite{zhang2018unreasonable}, which correlates more closely with human perceptual judgment.

\noindent\textbf{Baselines.} We quantitatively and qualitatively compare GO-PRE against a random selection baseline, established approaches including ActiveNeRF~\cite{pan2022activenerf} and FisherRF~\cite{jiang2023fisherrf}, and state-of-the-art active view selection methods such as GauSS-MI~\cite{xie2025gauss} and POp-GS~\cite{wilson2025pop}. For POp-GS, which introduces a family of optimality~\cite{kiefer1974general} strategies, we compare against its best-performing D-Optimality variant. To ensure a fair comparison with ActiveNeRF~\cite{pan2022activenerf}, we adopted the adaptation methodology proposed in FisherRF~\cite{jiang2023fisherrf} to implement its variance estimation algorithm using CUDA within the 3D Gaussian Splatting backend, effectively transplanting the next-best-view strategy into the 3D Gaussian Splatting~\cite{kerbl3Dgaussians} framework. For all comparative methods, we utilized the default hyperparameters provided in their respective papers or official open-source implementations.
\begin{figure*}[t]
  \centering
  \includegraphics[width=\textwidth]{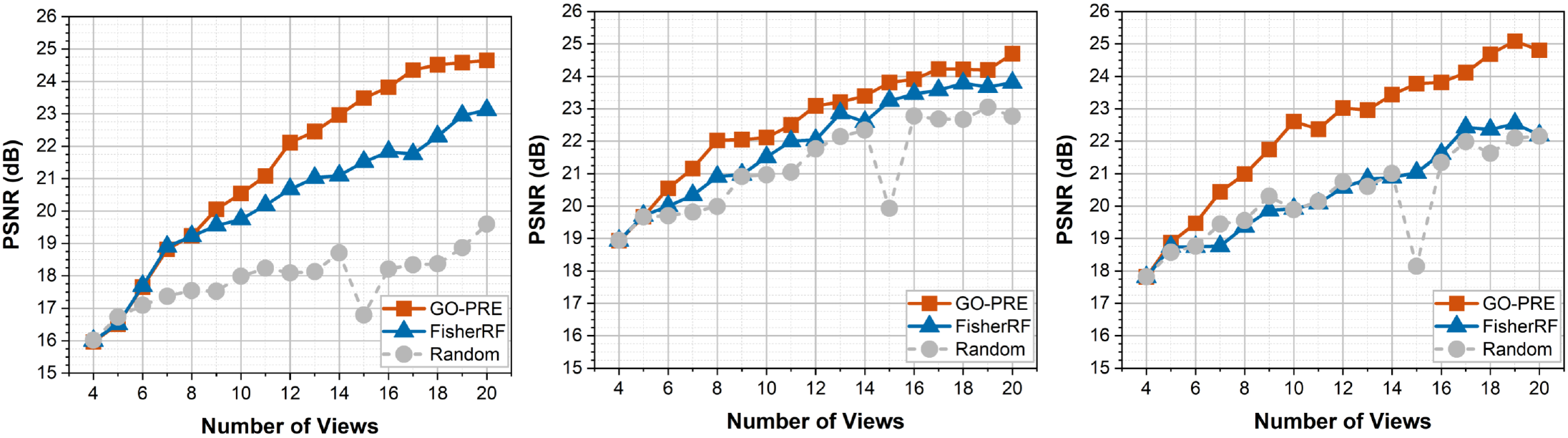}
  
  \hspace*{0.012\textwidth}%
  \makebox[0.335\textwidth][c]{\small (a) bonsai}
  \makebox[0.332\textwidth][c]{\small (b) Kitchen}
  \makebox[0.31\textwidth][c]{\small (c) Room}
  \caption{Reconstruction quality (PSNR) comparison on the Mip-NeRF360 dataset as new views are added.}
  \label{fig:active-view-selection-curves}
\end{figure*}

\noindent\textbf{Evaluation Protocols.} We design two experimental protocols to evaluate the versatility of our method across different reconstruction objectives.

\noindent\textit{Protocol I: Global Reconstruction (Table~\ref{tab:global-reconstruction}).}
This protocol follows the standard active view selection setting used in prior work, where the goal is to maximize reconstruction quality across the entire scene. For the Blender dataset, we utilize the official training split as our candidate pool and evaluate performance on the full test set (200 views per scene). For the Mip-NeRF360 and Tanks \& Temples datasets, we adopt the standard LLFF~\cite{mildenhall2019local} hold-out rule, selecting every 8th view for testing and using the remaining views as the candidate pool.

\noindent\textit{Protocol II: Goal-Oriented Reconstruction (Table~\ref{tab:goal-oriented-reconstruction}).}
To demonstrate the unique capability of GO-PRE in task-driven scenarios, we introduce a goal-oriented evaluation protocol. For each scene, we first compute the centroid of all camera poses and define a \emph{target view manifold} as a spatial sector formed by a $120^{\circ}$ clockwise rotation in the $x$--$y$ plane around this centroid. For the Blender dataset, we evaluate performance strictly on the subset of official test views located within this target manifold, using the complete training split as the candidate pool. For the Mip-NeRF360 and Tanks \& Temples datasets, we apply the LLFF~\cite{mildenhall2019local} hold-out rule to views within the target view manifold to construct the test set, while all remaining views serve as the candidate pool.

\noindent\textbf{Experimental Settings.} For all experiments, models are initialized with the same random seed and start from the same set of initial views. Each model is trained for a total of 30,000 iterations. To prevent degeneration during the incremental training process, we reset the opacity of the 3D Gaussians after each view selection step. Furthermore, all external training settings are kept identical across methods to ensure a fair comparison. We tailor the acquisition schedule to the target budget via an incremental selection scheme: for complex scenes such as Mip-NeRF360 and Tanks \& Temples, a Standard 20-View Schedule is employed where the model is initialized with 4 uniformly sampled views, followed by the addition of one new view every 100 epochs until a total of 20 views is reached. For simpler scenes in the Blender dataset and the sparsity stress test reported in Table~\ref{tab:sparse-view-reconstruction}, we adopt a stricter 2-view initialization, adding one new view every 100 epochs up to a 10-view limit. All experiments are conducted on a single NVIDIA RTX 3090 GPU. Under the default configuration, per-candidate evaluation in GO-PRE takes approximately 120\,ms on Mip-NeRF360 with 20 views.

To approximate the Goal Hessian Matrix $M$, at each view selection step, we randomly sample a set of $K$ probe poses $\{p_k\}$ from the pool of training views located within the user-specified target manifold. This sampling process is independent of the selection history. The sampling pool encompasses the entire training set for Global Reconstruction, while being restricted to the subset within the target sector for Goal-Oriented tasks. We set the sample size to a fixed $K=25$ for the Blender dataset, and to $\frac{1}{8}$ of the total view count for the Mip-NeRF360 and Tanks \& Temples datasets.

\noindent\textbf{Results.}
Quantitative results for active view selection are summarized in Table~\ref{tab:global-reconstruction}, Table~\ref{tab:goal-oriented-reconstruction}, and Table~\ref{tab:sparse-view-reconstruction}, while Figure~\ref{fig:active-view-selection-curves} visualizes the step-by-step performance gains. As demonstrated, GO-PRE consistently achieves superior performance across all evaluated metrics and datasets. Figure~\ref{fig:teaser} and Figure~\ref{fig:qualitative-results} provide qualitative comparisons on the Blender~\cite{mildenhall2021nerf}, Mip-NeRF360~\cite{barron2022mip}, and Tanks \& Temples~\cite{knapitsch2017tanks} datasets, visually demonstrating the enhancement in reconstruction fidelity. In the global reconstruction task (Table~\ref{tab:global-reconstruction}), GO-PRE outperforms information-theoretic baselines, validating the superiority of directly optimizing predictive rendering entropy over indirect information metrics. This advantage is further amplified in goal-oriented scenarios (Table~\ref{tab:goal-oriented-reconstruction}), where GO-PRE explicitly concentrates uncertainty reduction within the target sector, thereby achieving superior reconstruction quality within the target view manifold. The sparse-view stress test (Table~\ref{tab:sparse-view-reconstruction}) confirms our robustness; even under a tight 10-view budget, GO-PRE significantly surpasses baselines by identifying critical viewpoints early in the reconstruction process.

\begin{table}[t]
\caption{Results on Mip-NeRF360 Dataset with 10 Views.}
\label{tab:sparse-view-reconstruction}
\centering
\setlength{\tabcolsep}{6pt}
\renewcommand{\arraystretch}{1.15}

\begin{tabular}{c|ccc}
\toprule
Method & PSNR$\uparrow$ & SSIM$\uparrow$ & LPIPS$\downarrow$ \\
\midrule
Random    & 15.8774 & 0.4311 & 0.4682 \\
FisherRF  & \third{16.9470} & \third{0.4745} & 0.4445 \\
GauSS-MI  & 16.8920 & 0.4640 & \third{0.4440} \\
POp-GS    & \second{17.7380} & \second{0.5080} & \second{0.4124} \\
Ours      & \best{18.0650} & \best{0.5090} & \best{0.4105} \\
\bottomrule
\end{tabular}
\end{table}

\begin{figure}
    \centering
    \includegraphics[width=1\linewidth]{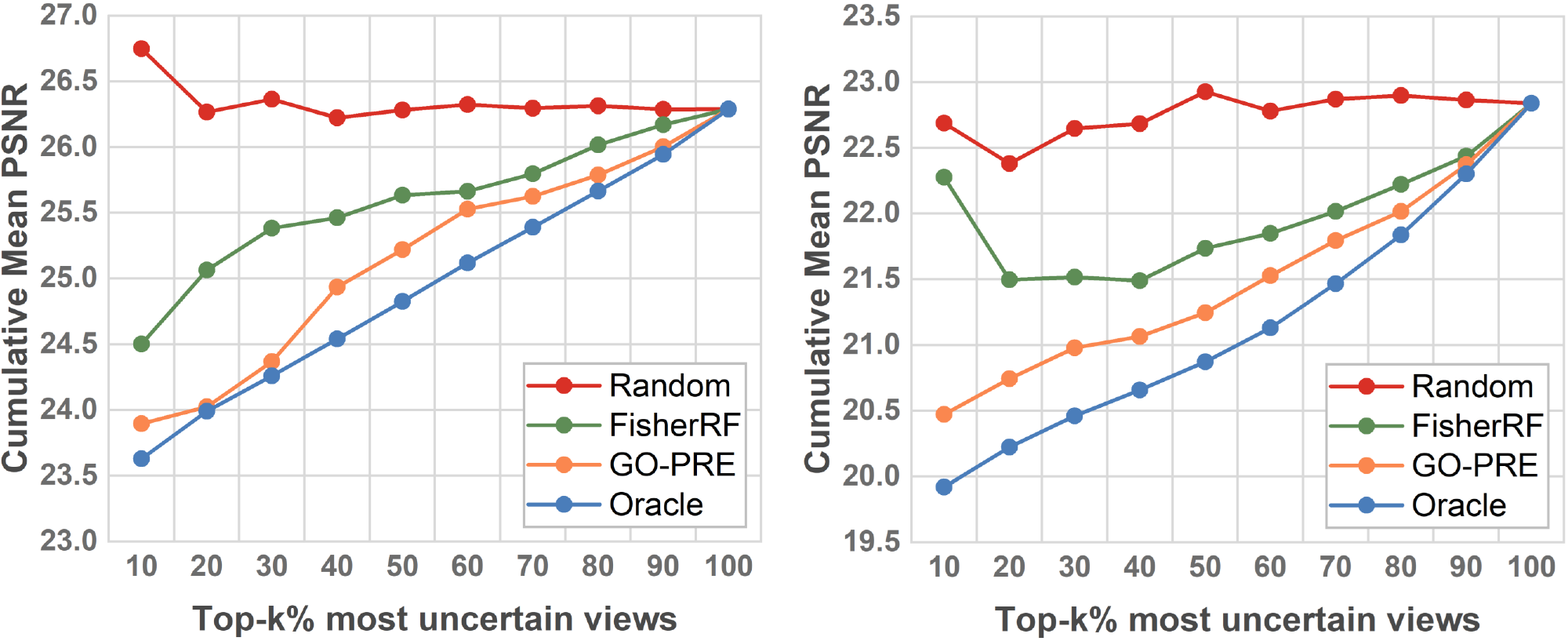}

    \makebox[0.48\linewidth][c]{\small (a) ficus}%
    \hspace*{0.05\linewidth}\makebox[0.45\linewidth][c]{\small (b) lego}
    \caption{Correlation between predicted uncertainty and actual reconstruction quality on the Blender dataset.}
    \label{fig:uncertainty-sparsification}
\end{figure}

\begin{table}[t]
\caption{Ablation study of key components on the Mip-NeRF360 Dataset.}
\label{tab:ablation}
\centering
\setlength{\tabcolsep}{6pt}
\renewcommand{\arraystretch}{1.15}
\begin{tabular}{c|ccc}
\toprule
Method & PSNR$\uparrow$ & SSIM$\uparrow$ & LPIPS$\downarrow$ \\
\midrule
Goal--Metric Misalignment & 19.295           & 0.579            & 0.380 \\
w/o Goal Hessian          & 20.532           & 0.613            & 0.351 \\
Trace Approximation       & \third{20.632}   & 0.604            & 0.357 \\
w/o Observation Noise     & 20.603           & \third{0.618}    & \third{0.349} \\
Direct-MC Sampling        & \second{21.013}  & \second{0.625}   & \second{0.341} \\
Ours                      & \best{21.228}    & \best{0.632}     & \best{0.337} \\
\bottomrule
\end{tabular}
\end{table}

\subsection{Uncertainty Quantification}

As discussed in Sec.~\ref{sec:uncertainty-quantification}, our framework naturally extends to quantify predictive uncertainty for unseen views. To evaluate the correlation between uncertainty estimates and actual rendering errors, we employ sparsification plots~\cite{ilg2018uncertainty}, where test views are sorted by predicted uncertainty and the cumulative mean reconstruction quality is plotted in decile increments, as visualised in Figure~\ref{fig:uncertainty-sparsification}. Since uncertainty and reconstruction quality are inversely related, a well-calibrated estimator should rank the worst-reconstructed views as most uncertain, yielding a curve that rises from left to right. We train a 3DGS~\cite{kerbl3Dgaussians} model on a randomly selected subset of 10 images for 2,000 iterations. Ensuring strict consistency, each method ranks the test views based on its predicted uncertainty using the identical random seed and fixed trained model.

In Figure~\ref{fig:uncertainty-sparsification}, we introduce an Oracle baseline which ranks views based on their ground truth rendering PSNR, representing the ideal error-uncertainty alignment, alongside a Random baseline that sorts views arbitrarily. As illustrated, the sparsification curve generated by GO-PRE aligns significantly closer to the Oracle curve compared to both FisherRF~\cite{jiang2023fisherrf} and the Random baseline. This proximity demonstrates that our predictive entropy provides a much more accurate estimation of potential rendering errors, effectively differentiating between reliable and unreliable viewpoints better than parameter-space baselines.

\subsection{Analysis}

Finally, we conduct ablation and sensitivity analyses to investigate the contribution of key components within our information gain quantification framework and to characterize the framework's behavior under varied configurations. All experiments are conducted on the Mip-NeRF360~\cite{barron2022mip} dataset with 20 views, and quantitative results are summarized in Tables~\ref{tab:ablation}--\ref{tab:candidate-pool}, with each component analyzed in detail below.

\noindent\textbf{w/o Goal Hessian ($\mathbf{M} = \mathbf{I}$).}
We replace the computed Goal Hessian matrix $\mathbf{M}$ with the identity matrix $\mathbf{I}$, removing the probe-derived guidance and treating all parameter directions as equally relevant to the target manifold. As reported in Table~\ref{tab:ablation}, this variant underperforms our full model by 0.70\,dB PSNR. The drop confirms that $\mathbf{M}$ carries the goal-orienting weighting that aligns view selection with the user-specified target manifold.

\noindent\textbf{w/o Observation Noise.}
The additive Gaussian noise term $\eta$ in Eq.~\ref{eq:predictive-model} models the aleatoric uncertainty of pixel observations, with its variance $\tau^2$ entering the acquisition score as a precision-scale parameter. We implement this ablation by taking the limit $\tau \to 0$, under which the Goal Hessian loses its effect on the candidate ranking. As reported in Table~\ref{tab:ablation}, this variant underperforms our full model by 0.63\,dB PSNR, confirming that $\tau^2$ plays a meaningful role in shaping the acquisition signal.

\noindent\textbf{Goal--Metric Misalignment.}
To verify the goal-orienting capability of our method, we deliberately misalign the probe set with the evaluation set: probes are drawn from the $120^{\circ}$ target view manifold defined in Sec.~\ref{sec:active-view-selection}, while reconstruction quality is evaluated on the full $360^{\circ}$ scene. Under this mismatch, this variant exhibits a PSNR drop of 1.93\,dB in Table~\ref{tab:ablation}, the largest among all ablations.

\noindent\textbf{Trace Approximation.}
To isolate the efficacy of our entropic objective, we benchmark it against a trace-based formulation~\cite{jiang2023fisherrf}. Premised on the approximation $\log \det(I + \mathbf{A}) \approx \text{Tr}(\mathbf{A})$, which is analytically valid when the spectral radius is small, we employ this trace metric as a surrogate for our information gain objective. Table~\ref{tab:ablation} demonstrates that this strategy performs significantly worse than our method. We attribute this to the fact that the trace-based computation of information gain fails to capture the diminishing marginal utility of information. Consequently, the policy tends to over-prioritize views with large gradient norms regardless of the actual uncertainty, resulting in redundant sampling and reduced reconstruction efficiency.

\textbf{Direct-MC Sampling.} To examine the practical validity of our Jensen-based surrogate, we replace it with a Monte Carlo estimator that bypasses Jensen's inequality and directly computes the expected predictive uncertainty in Eq.~(7) under the same probe budget. As shown in Table~\ref{tab:ablation}, Direct-MC already ranks second across all three metrics---surpassing every other ablation variant---yet GO-PRE further improves over it by 0.22\,dB PSNR while requiring 28\% less peak GPU memory (5.47 vs.\ 7.58\,GiB) and 11\% less per-candidate time (120.18 vs.\ 135.06\,ms). A plausible explanation lies in the order of operations: GO-PRE aggregates Fisher information across probes into the Goal Hessian $\mathbf{M}$ before evaluating the log-determinant, whereas Direct-MC computes a log-determinant per probe and averages afterwards---an ordering likely more susceptible to Monte Carlo variance.

\textbf{Sensitivity to Probe Count $K$.} Table~\ref{tab:probe-count} examines the impact of the probe count $K$ used to estimate the Goal Hessian $\mathbf{M}$. Performance saturates around $K = N/8$, with $K = N/4$ yielding no statistically meaningful improvement, validating our default choice. As $K$ decreases, both PSNR and per-candidate runtime degrade gracefully, with no pathological transition. Remarkably, even at $K = N/64$---roughly 3.3 probes per round on Mip-NeRF360---GO-PRE achieves 20.87\,dB PSNR, still surpassing the runner-up POp-GS~\cite{wilson2025pop} (20.62\,dB in Table~\ref{tab:global-reconstruction}). This robustness is consistent with the variance-damping picture suggested by the Direct-MC comparison above, and makes GO-PRE readily adaptable to resource-constrained deployment, where users can tune $K$ to trade modest quality for substantial runtime savings.

\begin{table}[t]
\caption{Sensitivity to the number of probes $K$ on the Mip-NeRF360 Dataset with 20 views. Per-cand. denotes the per-candidate selection time.}
\label{tab:probe-count}
\centering
\setlength{\tabcolsep}{6pt}
\renewcommand{\arraystretch}{1.15}
\begin{tabular}{c|cc}
\toprule
$K$ & PSNR (mean$\pm$std) $\uparrow$ & Per-cand. (ms) $\downarrow$ \\
\midrule
$N/4$            & \best{21.24 $\pm$ 0.07}   & 135.77 \\
$N/8$ (default)  & \second{21.23 $\pm$ 0.09} & 120.18 \\
$N/12$           & \third{21.14 $\pm$ 0.12}  & 116.17 \\
$N/16$           & 21.10 $\pm$ 0.13          & \third{114.06} \\
$N/32$           & 20.98 $\pm$ 0.11          & \second{109.69} \\
$N/64$           & 20.87 $\pm$ 0.18          & \best{107.85} \\
\bottomrule
\end{tabular}
\end{table}

\begin{table}[t]
\caption{Sensitivity to random candidate-pool reduction on the Mip-NeRF360 Dataset with 20 views. Selection time denotes the average per-round selection time.}
\label{tab:candidate-pool}
\centering
\setlength{\tabcolsep}{6pt}
\renewcommand{\arraystretch}{1.15}
\begin{tabular}{c|cc}
\toprule
Reduction & PSNR (mean$\pm$std)$\uparrow$ & Time (s)$\downarrow$ \\
\midrule
None (default) & \best{21.23 $\pm$ 0.09}   & 22.23 \\
25\%           & \second{21.12 $\pm$ 0.11} & \third{17.85} \\
33\%           & \third{21.04 $\pm$ 0.17}  & \second{16.31} \\
50\%           & 20.93 $\pm$ 0.15          & \best{13.25} \\
\bottomrule
\end{tabular}
\end{table}

\noindent\textbf{Robustness to Candidate Pool Reduction.}
Since the per-step computational cost grows linearly with the candidate pool size, we conduct an ablation on the size of the candidate view pool to study how the framework can be adapted in practice for deployment on resource-constrained edge devices or latency-sensitive applications. Specifically, we randomly remove a fraction of the candidate views at each selection step, and report both the average per-round selection time and the reconstruction quality. As shown in Table~\ref{tab:candidate-pool}, PSNR degrades only gracefully even under substantial pool reduction: with half of the candidate views removed, the performance loss is bounded within 0.30\,dB---still outperforming all competing baselines in Table~\ref{tab:global-reconstruction}---while the selection time is reduced by roughly 40\%. Pool reduction thus provides an additional axis along which GO-PRE can be adapted to resource-constrained deployment.

\section{Conclusion}
In this paper, we presented GO-PRE, a principled active view selection framework that maximizes predictive rendering entropy reduction over a user-specified target view manifold, aligning view acquisition with the rendered outputs that ultimately matter. Across diverse benchmarks, GO-PRE consistently improves reconstruction quality and yields more reliable uncertainty estimates than state-of-the-art baselines, while enabling interactive goal specification. GO-PRE establishes prediction-space information gain as a principled foundation for active 3D reconstruction, moving beyond surrogate parameter-space and geometric heuristics.

\section{Limitations and Future Work}
In real-world active reconstruction, cameras navigate a continuous pose space, yet existing methods---including GO-PRE---still select the next view from a discrete candidate pool. The GO-PRE acquisition score is formally differentiable with respect to camera pose and could in principle support gradient-based optimization over continuous poses, but two structural barriers stand in the way. First, the score depends on pose through the rendering Jacobian, so back-propagating it requires  mixed second-order derivatives of the renderer, which are prohibitive for the high-dimensional parameters of 3D Gaussian Splatting. Second, the rasterizer is only piecewise smooth, as depth sorting and visibility transitions introduce discontinuities that destabilize local linearizations. We regard the design of acquisition objectives compatible with continuous, gradient-based pose optimization as a promising direction for truly continuous next best view selection.

\newpage

\section*{Acknowledgements}
This work was supported by the National Natural Science Foundation of China (Nos. 62576109 and 62406075), the Scientific and Technological Innovation Action Plan of Shanghai Science and Technology Committee (No. 25511104402), the National Key Research and Development Program of China (No. 2023YFC3604802), and the Shanghai Key Technology R\&D Program (Grant No. 25511107200).

\section*{Impact Statement}
This paper presents GO-PRE, a framework designed to advance the field of active 3D reconstruction and uncertainty quantification. By enabling more efficient and goal-oriented view selection, our work has the potential to significantly enhance the capabilities of autonomous systems, such as search-and-rescue robots exploring unknown environments and drones performing precision infrastructure inspections. These advancements can lead to reduced energy consumption and improved operational safety in various real-world applications. While the improved fidelity of 3D reconstruction could theoretically be misused in surveillance contexts, the primary focus of this research is on the fundamental algorithmic improvement of information gain quantification, which we believe contributes positively to the development of reliable and interpretable machine learning systems.

\bibliography{references}
\bibliographystyle{icml2026}

\newpage
\appendix
\onecolumn
\section{Appendix}

\subsection{Detailed Mathematical Derivations}
\label{sec:appendix-derivations}

In this section, we provide the step-by-step derivation of the GO-PRE Score presented in Sec~\ref{sec:go-pre-score} of the main paper.

\subsubsection{Derivation of Equation (7)}
Recall that the posterior covariance update after observing a new view $x$ is given by:
\begin{equation}
    \Sigma_{x} = (\Sigma^{-1} + \tau^{-2} J_x^\top J_x)^{-1}.
\end{equation}
The rendering uncertainty at a target pose $p$ is quantified by the log-determinant of the predictive covariance $\Lambda_p = J_p \Sigma_x J_p^\top + \tau^2 I$. Using the \textbf{Matrix Determinant Lemma}, specifically the identity $\det(A + UV^\top) = \det(I + V^\top A^{-1} U) \det(A)$, we can rewrite the determinant term. Let $A = \tau^2 I$, $U = J_p \Sigma_x^{1/2}$, and $V = \Sigma_x^{1/2} J_p^\top$. Note that we perform the eigen-decomposition on $\Sigma_x$ such that $\Sigma_x = \Sigma_x^{1/2} \Sigma_x^{1/2}$.

The predictive entropy term can be transformed as:
\begin{equation}
\begin{aligned}
    \log \det(J_p \Sigma_x J_p^\top + \tau^2 I) &= \log \det(\tau^2 I + (J_p \Sigma_x^{1/2})(\Sigma_x^{1/2} J_p^\top)) \\
    &= \log \left( \det(\tau^2 I) \cdot \det(I + \tau^{-2} (\Sigma_x^{1/2} J_p^\top) (J_p \Sigma_x^{1/2})) \right) \\
    &= C + \log \det(I + \tau^{-2} \Sigma_x^{1/2} (J_p^\top J_p) \Sigma_x^{1/2}),
\end{aligned}
\end{equation}
where $C$ is a constant related to the noise variance $\tau^2$. By defining the Fisher Information at pose $p$ as $S(p) = J_p^\top J_p$, we recover Eq.~(\ref{eq:parameter-space-uncertainty}) from the main text:
\begin{equation}
    \mathcal{U}_x \equiv \mathbb{E}_{p \sim q} \left[ \frac{1}{2} \log \det(I + \tau^{-2} \Sigma_x^{1/2} S(p) \Sigma_x^{1/2}) \right].
\end{equation}

\subsubsection{Derivation of the GO-PRE Score via Jensen's Inequality}
The objective is to minimize the expected uncertainty $\mathcal{U}_x$. Since the function $f(A) = \log \det(A)$ is concave over the cone of positive-definite matrices, applying Jensen's Inequality ($\mathbb{E}[f(X)] \leq f(\mathbb{E}[X])$) gives us an upper bound on the uncertainty:
\begin{equation}
\begin{aligned}
    \mathcal{U}_x &= \mathbb{E}_{p \sim q} \left[ \frac{1}{2} \log \det(I + \tau^{-2} \Sigma_x^{1/2} S(p) \Sigma_x^{1/2}) \right] \\
    &\leq \frac{1}{2} \log \det \left( \mathbb{E}_{p \sim q} [I + \tau^{-2} \Sigma_x^{1/2} S(p) \Sigma_x^{1/2}] \right) \\
    &= \frac{1}{2} \log \det \left( I + \tau^{-2} \Sigma_x^{1/2} \mathbb{E}_{p \sim q}[S(p)] \Sigma_x^{1/2} \right).
\end{aligned}
\end{equation}
By defining the \textbf{Goal Hessian Matrix} $M = \mathbb{E}_{p \sim q}[S(p)]$ and setting $\tau=1$ for simplicity (as is standard in prior works), we arrive at the GO-PRE Score:
\begin{equation}
    Score_{GO-PRE}(x) = \log \det (I + \Sigma_x^{1/2} M \Sigma_x^{1/2}).
\end{equation}
Minimizing this upper bound effectively pushes down the actual predictive entropy over the target view manifold.

\end{document}